\documentclass[letterpaper, 10 pt, conference]{ieeeconf}  

\IEEEoverridecommandlockouts                              

\usepackage{caption}
\usepackage{subcaption}
\usepackage{graphicx}
\usepackage{tikz}
\usepackage{url} 
\usetikzlibrary{calc}
\usetikzlibrary{arrows.meta}
\usetikzlibrary{positioning}
\usetikzlibrary{fit}
\usetikzlibrary{shapes}
\usetikzlibrary{shapes.geometric, arrows}
\tikzstyle{startstop} = [rectangle, rounded corners, minimum width=3cm, minimum height=1cm,text centered, draw=black, fill=white] 
\tikzstyle{io} = [trapezium, trapezium left angle=80, trapezium right angle=100, minimum width=1.5cm, minimum height=1cm, text centered, draw=black, fill=white]
\tikzstyle{process} = [rectangle, minimum width=3cm, minimum height=1cm, text centered, draw=black, fill=white]
\tikzstyle{decision} = [diamond, minimum width=3cm, minimum height=1cm, text centered, draw=black, fill=white, aspect=2] 
\tikzstyle{title} = [rectangle, minimum width=3cm, minimum height=1cm, text centered, draw=white, fill=white]
\tikzstyle{arrow} = [thick,->,>=stealth]
\usepackage{array}
\usepackage{cite}
\usepackage{amsmath,amssymb,amsfonts}
\usepackage{algorithmic}
\usepackage{textcomp}
\usepackage{xcolor}
\def\BibTeX{{\rm B\kern-.05em{\sc i\kern-.025em b}\kern-.08em
    T\kern-.1667em\lower.7ex\hbox{E}\kern-.125emX}}
\newcommand\SmallMatrix[1]{{%
  \small\arraycolsep=\arraycolsep\ensuremath{\begin{bmatrix}#1\end{bmatrix}}}}

\title{\LARGE \bf
A Novel Methodology for Autonomous Planetary Exploration \\ Using Multi-Robot Teams
}

\author{Sarah Swinton*$^{1}$, Jan-Hendrik Ewers$^{1}$, Euan McGookin$^{1}$, David Anderson$^{1}$, and Douglas Thomson$^{1}$
\thanks{This work has been submitted to the IEEE for possible publication. Copyright may be transferred without notice, after which this version may no longer be accessible.}
\thanks{This document is the result of a research project funded by the Engineering and Physical Sciences Research Council.}
\thanks{* Corresponding Author}
\thanks{$^{1}$ Sarah Swinton, Jan-Hendrik Ewers, Euan McGookin, David Anderson, and Douglas Thomson are with the James Watt School of Engineering,
        University of Glasgow, Glasgow, United Kingdom. Email:
        {s.swinton.1@research.gla.ac.uk; j.ewers.1@research.gla.ac.uk; euan.mcgookin@glasgow.ac.uk; dave.anderson@glasgow.ac.uk; douglas.thomson@glasgow.ac.uk}}, 
}

\begin{document}

\maketitle
\thispagestyle{empty}
\pagestyle{empty}

\begin{abstract}
One of the fundamental limiting factors in planetary exploration is the autonomous capabilities of planetary exploration rovers.
This study proposes a novel methodology for trustworthy autonomous multi-robot teams which incorporates data from multiple sources (HiRISE orbiter imaging, probability distribution maps, and on-board rover sensors) to find efficient exploration routes in Jezero crater.  A map  is generated, consisting of a 3D terrain model, traversability analysis, and probability distribution map of points of scientific interest. A three-stage mission planner generates an efficient route, which maximises the accumulated probability of identifying points of interest. A 4D RRT* algorithm is used to determine smooth, flat paths, and prioritised planning is used to coordinate a safe set of paths. The above methodology is shown to coordinate safe and efficient rover paths, which ensure the rovers remain within their nominal pitch and roll limits throughout operation.
\end{abstract}

\vspace*{-3mm}
\section{INTRODUCTION}
Trustworthy operation of mobile robotic systems depends on their safe navigation of the environment around them. Primarily this depends on their guidance system and in particular the associated path or trajectory planning algorithm. Nowhere is this more evident than in the field of planetary exploration, where damage to the rover's hardware could mean the end of a mission. 
One approach to maximising the temporal functionality of a planetary exploration mission is to utilise a team of small robots, in place of one very large robot. NASA's Mars 2020 Mission has been the first planetary exploration mission to utilise two robotic platforms working in close proximity: the Perseverance Rover, and the Ingenuity Mars Helicopter \cite{farley2020}. 
Further research is currently being carried out to evaluate the efficacy of multi-robot teams for both surface \cite{fong2021}\cite{swinton2022novel} and cave exploration \cite{fink2015}\cite{vaquero2018}. 

A key consideration for multi-robot planetary exploration is the increased complexity of the entire system compared to a single robot. As system complexity increases, so too does the workload and stress of a human operator. Therefore, high levels of autonomy within multi-robot systems are essential to reduce the required cognitive load on human operators \cite{stOnge2019}. Early planetary rover missions employed very little autonomy in terms of path planning, relying instead on driving commands from human operators \cite{bajracharya2008}. Each subsequent rover has utilised increased levels of autonomy, with the most recent, the Perseverance rover, setting records for the longest distance driven on a single sol (347.7m) and the longest distance driven without human review (699.9m) \cite{verma2023}.

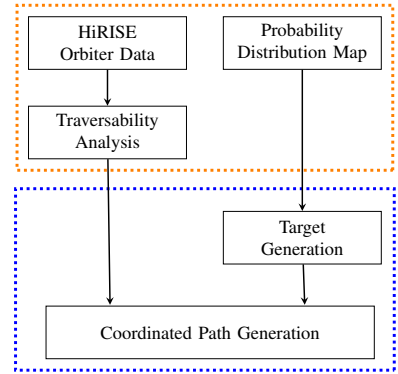
\begin{figure}[h]
    \centering
    \scalebox{0.7}{
    \begin{tikzpicture}[node distance=1.7cm]
        \node (pro01) [process,yshift = 0.85cm]{\parbox{2.6cm}{\centering HiRISE\\ Orbiter Data}}; 
        \node (pro02) [process, right of=pro01,xshift = 2cm]{\parbox{2.6cm}{\centering Probability\\ Distribution Map}};
        \node (pro11) [process, below of=pro01]{\parbox{2.6cm}{\centering Traversability\\ Analysis}}; 
        \draw[dotted,orange,ultra thick] ($(current bounding box.south east)+(6pt,-6pt)$) rectangle ($(current bounding box.north west)+(-6pt,6pt)$);

        \node (pro03) [process, below of=pro11,xshift = 1.95cm,yshift = -2.1cm]{\parbox{6cm}{\centering Coordinated Path Generation}}; 
        \node (pro12) [process, below of=pro02, yshift = -2cm]{\parbox{2.6cm}{\centering Target\\ Generation}}; 
        \draw[dotted,blue,ultra thick] ($(current bounding box.south east)+(0pt,-6pt)$) rectangle ($(current bounding box.north west)+(-0pt,-100pt)$);

        \node[above of=pro03, xshift = -1.9cm,yshift = -1.3cm]  (pointA1) {};    
        \node[left of=pointA1]  (pointA2) {};    
        \node[above of=pro03,xshift = 1.8cm,yshift = -1.3cm]  (pointB1) {};    
        
        \draw [arrow] (pro01) -- (pro11);
        \draw [arrow] (pro02) -- (pro12);
        \draw [arrow] (pro11) -- (pointA1);
        \draw [arrow] (pro12) -- (pointB1);
    \end{tikzpicture}
    } 
    \caption{Autonomous exploration system architecture}
    \label{fig:architecture}
\end{figure}

These more recent missions have demonstrated the benefit of employing autonomy. The work presented in this paper follows on in this theme by proposing a novel methodology for autonomous exploration using a team of wheeled rovers. This involves the trajectory coordination of multiple rovers as they explore the planetary surface and perform a designated task.
Fig. \ref{fig:architecture} shows the architecture of the proposed methodology. First, an environment model is created (outlined in orange), which takes into account the traversability of terrain, and the location of points of scientific interest using a probability distribution map. 
Next, mission planning is carried out (outlined in blue). This consists of the generation of an efficient set of targets within the environment, and evaluation of coordinated rover paths.

\vspace*{-2mm}
The contributions of this work are: 
\begin{itemize}
    \item a mapping approach for planetary exploration mission sites that combines data on terrain traversability and the location of points of scientific interest
    \item a novel approach to autonomous coordinated exploration for a team of planetary exploration rovers in a 3D environment.
\end{itemize}

This paper presents this work in the following manner. Section \ref{sec:rover} sets out the multi-rover system. The mapping methodology is discussed in Section \ref{sec:map}. Section \ref{sec:missionPlanning} defines the mission planning approach.
Section \ref{sec:results} describes the experiments carried out to evaluate the proposed methods. Finally, Section \ref{sec:conclusions} summarises the outcomes from this study and the overall conclusions that can be drawn from this work. 
\newpage
\section{MULTI-ROVER SYSTEM} \label{sec:rover}
Planetary exploration missions are subject to strict financial and payload constraints \cite{planetarySociety}. It would not be feasible to plan a multi-rover mission where each individual rover has the technical capabilities of, for example, the Perseverance rover, and in turn the same constraints. A multi-rover team should be able to be stowed in a launch capsule that has regularly been used to hold a single, larger rover. Further, a reduced sensor suite may have to be employed. However, the members of a multi-rover team could be given specialised roles, each of which has a specialised and complementary set of sensors, to allow the team to carry out complex exploration tasks in a similar way to existing single rover systems. These specialised roles could include scouts, drillers, image analysers, and sample storers/carriers. It is important to note that reduced individual capability should not come at the cost of overall mobility as planetary exploration rovers (PERs) must be able to traverse uneven terrain and slopes.

\begin{figure}[htbp]
    \centering
    \includegraphics[width=0.7\linewidth]{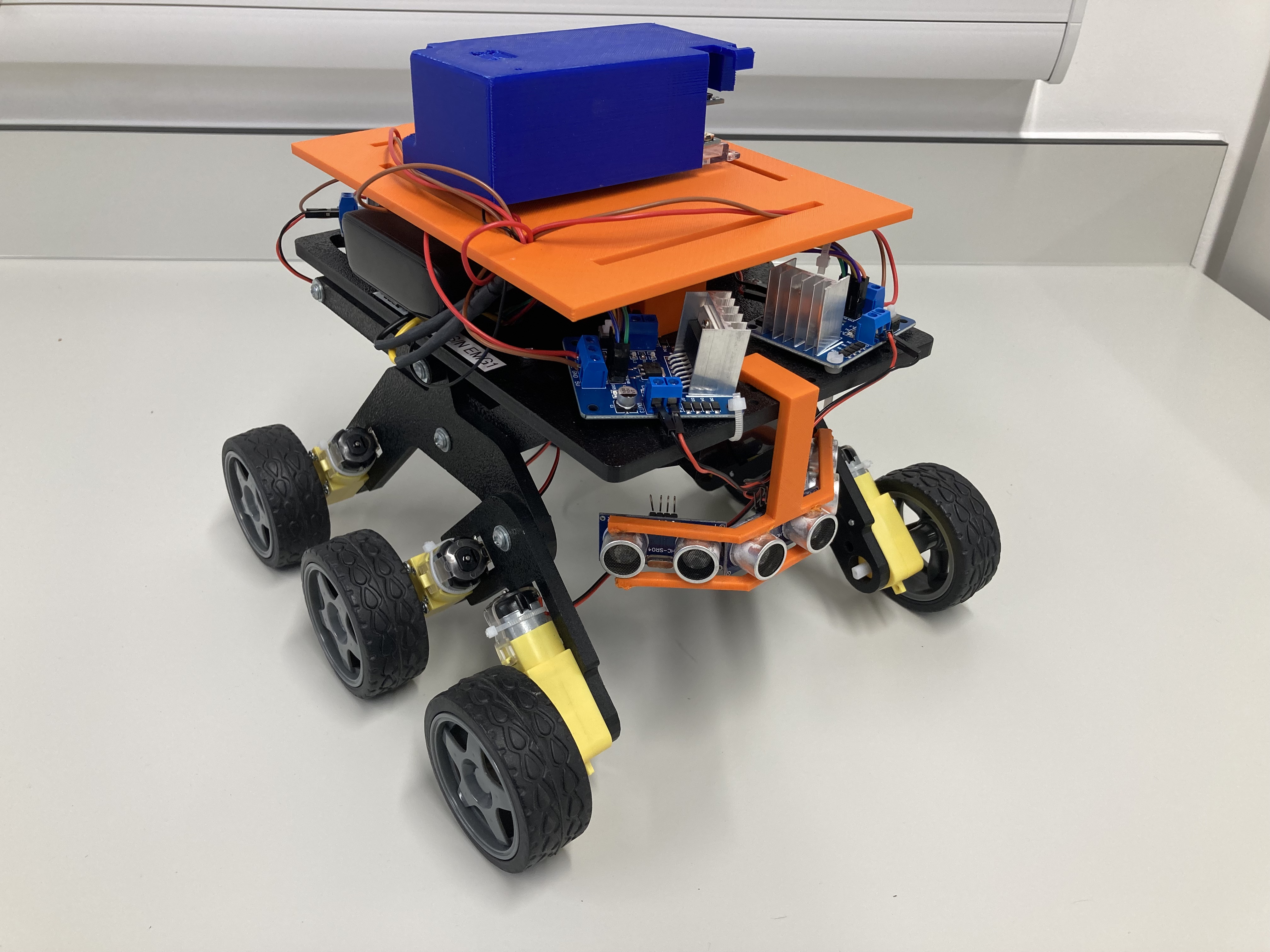}
    \caption{Rocker bogie runt rAover} \label{fig:rbrRover}
\end{figure}

To satisfy the requirements outlined above, a bogie runt rover \cite{servoCityRBR} has been selected for this work (Fig. \ref{fig:rbrRover}). This rover has a small form factor ($0.271$m$\times$$0.251$m$\times$$0.144$m), and a six-wheel rocker bogie suspension in line with the baseline mobility characteristics of current PERs \cite{flessa2014}.
The multi-rover team consists of five rocker bogie runt rovers, which have been modelled mathematically and simulated in MATLAB.

\vspace*{-3mm}
The rover's rigid body dynamics, with reference to the body-fixed frame and Mars-fixed frame, can be described by the matrix relationships shown in Equations (\ref{eqn:eqOfMotionGen}, \ref{eqn:eqOfMotionComp1}, \ref{eqn:eqOfMotionComp2}) \cite{fossen1994}. 

\renewcommand\arraystretch{1.5}
\begin{equation}
    \SmallMatrix{\boldsymbol{\dot{v}}\\\boldsymbol{\dot{\eta}}}
  = 
  \SmallMatrix{\boldsymbol{\alpha}(v) & \boldsymbol{\beta}(v)\\ \boldsymbol{J}(\eta) & 0}
  \SmallMatrix{\boldsymbol{v} \\ \boldsymbol{\eta}}
  +
  \SmallMatrix{ -\boldsymbol{M}^{-1} \\0 } \boldsymbol{\tau}
  \label{eqn:eqOfMotionGen}
\end{equation} 
\begin{equation}
    \boldsymbol{\alpha} (v)= -(\boldsymbol{C}(v)+\boldsymbol{D}(v)) \boldsymbol{M}^{-1}
  \label{eqn:eqOfMotionComp1}
\end{equation}
\begin{equation}
    \boldsymbol{\beta} (v) = -\boldsymbol{g}(v) \boldsymbol{M}^{-1}
  \label{eqn:eqOfMotionComp2}
\end{equation}

Here, $\boldsymbol{v}$ is the body-fixed velocity vector and $\boldsymbol{\eta}$ is the inertially fixed position/orientation vector. $\boldsymbol{M}$ is the mass and inertia matrix, $\boldsymbol{C}(v)$ is the Coriolis  matrix, $\boldsymbol{D}(v)$ is the damping  matrix, $\boldsymbol{g}(v)$ represents  the gravitational  forces  and  moments, $\boldsymbol{J}(\eta)$ is an Euler matrix representing the trigonometric transformation from the body fixed reference frame to the Mars fixed reference frame, and the $\boldsymbol{\tau}$ vector  represents the  forces and moments generated by the actuators.

The rover utilises a line-of-sight algorithm to navigate towards its current target point. The control system consists of PID controllers for heading and velocity, respectively.
\section{ENVIRONMENT MAPPING} \label{sec:map}
To generate a route for the rover team that allows efficient exploration, a multi-layer environment model is generated. This process is set out in the following subsections. First, a 3D terrain model of the mission site is generated using orbiter data. Next, a traversability analysis is carried out on the 3D terrain model. Finally, a \textit{probability distribution map} (PDM) is defined, which sets out the probability of finding a scientific \textit{point of interest} (POI) at any given position.
\subsection{3D Terrain Model}
\begin{figure}[htbp]
    \centering
    \includegraphics[width=0.85\linewidth]{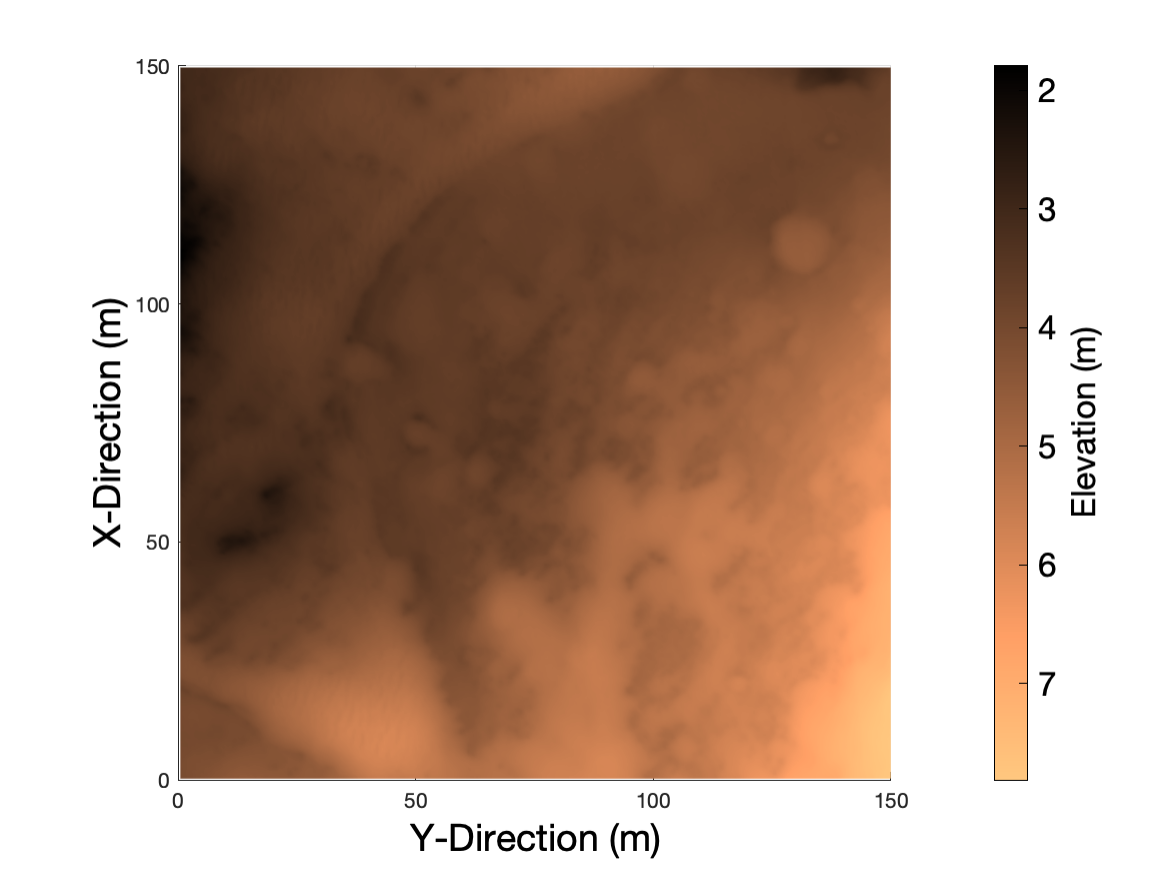}
    \caption{3D terrain model of the selected mission site} \label{fig:3dTerrainModel}
\end{figure}

In 2006, the Mars Reconnaissance Orbiter began operation surveying the surface of Mars \cite{zurek2007}. One of the primary instruments on the orbiter is the High-Resolution Imaging Science Experiment (HiRISE). HiRISE captures images of the surface of Mars with resolution of $\sim30$cm per pixel (from an altitude of $300$km) \cite{hiriseSpecs2012}. Using this data, a 3D terrain map of areas on the surface of Mars can be generated as a matrix of latitude, longitude, and elevation points. For this work, a $1500$m$\times$$1500$m mission site has been selected from within the Jezero crater. As the rover used in this work is approx. $1/10$th the size of the Perseverance rover, the environment model is scaled to $150$m$\times$$150$m. 
Within the simulation, the Martian surface is composed of a $600\times600$ block grid, where each block is $0.25$m$\times$$0.25$m. 
Fig. \ref{fig:3dTerrainModel} shows the 3D terrain model that has been simulated in MATLAB using HiRISE digital terrain modelling. 
\vspace*{-3mm}

\vspace*{-5mm}
\subsection{Traversability Analysis}
\begin{figure}[htbp]
    \centering
    \includegraphics[width=0.85\linewidth]{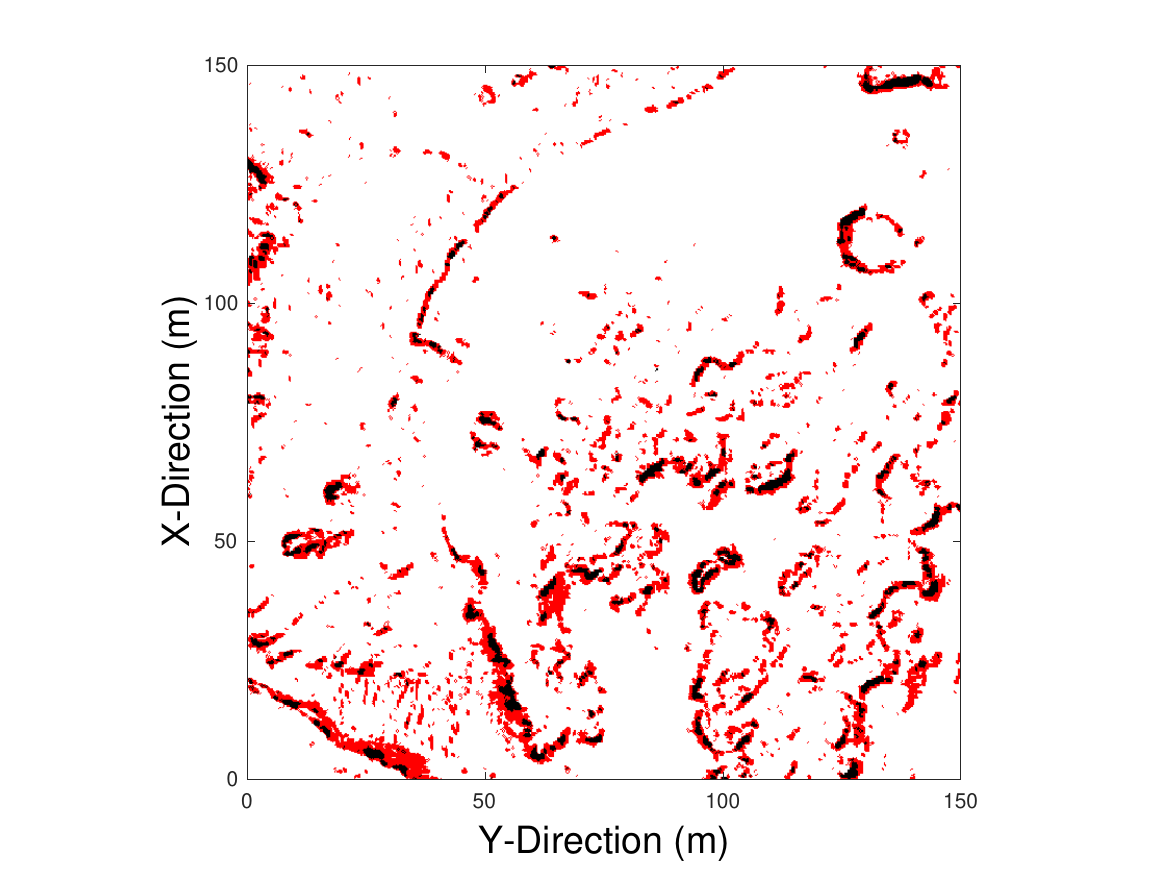}
    \caption{Traversability analysis of the selected mission site. Traversable terrain is shown in white, high risk terrain is shown in red, and impassable terrain is shown in black} \label{fig:travAnalysis}
\end{figure}

Planetary exploration rovers are subject to slip as they traverse steep slopes and rough terrain. To reduce the number of potential collisions, smooth and flat terrain should be explored when possible. A traversability analysis is carried out on the 3D terrain model where-by the elevation of neighbouring blocks in the $600\times600$ block grid are compared to find the slope angle required to traverse between the blocks. A given block inherits the worst-case slope angle. Nominal pitch and roll limits of $15^{\circ}$, in line with the nominal operational limits of the Perseverance rover \cite{rankin2020}, have been implemented. The traversability analysis determines which regions of each map are traversable, high risk, and impassable. Traversable terrain is safe to explore. High-risk terrain has a slope of $\theta >= 10^{\circ}$. Impassable terrain has a slope of  $\theta >= 15^{\circ}$. Fig. \ref{fig:travAnalysis} shows the resulting traversability map for the selected mission site. 
Here $89.50$\% of the map is traversable, 8.36\% of the map is high risk, and $2.14$\% of the map is impassable. 
\vspace*{-4mm}
\vspace*{-5mm}

\subsection{Probability Distribution Map}
\begin{figure}[htbp]
    \centering
     \includegraphics[clip, trim=3cm 0cm 0.9cm 0cm, width=0.75\linewidth]{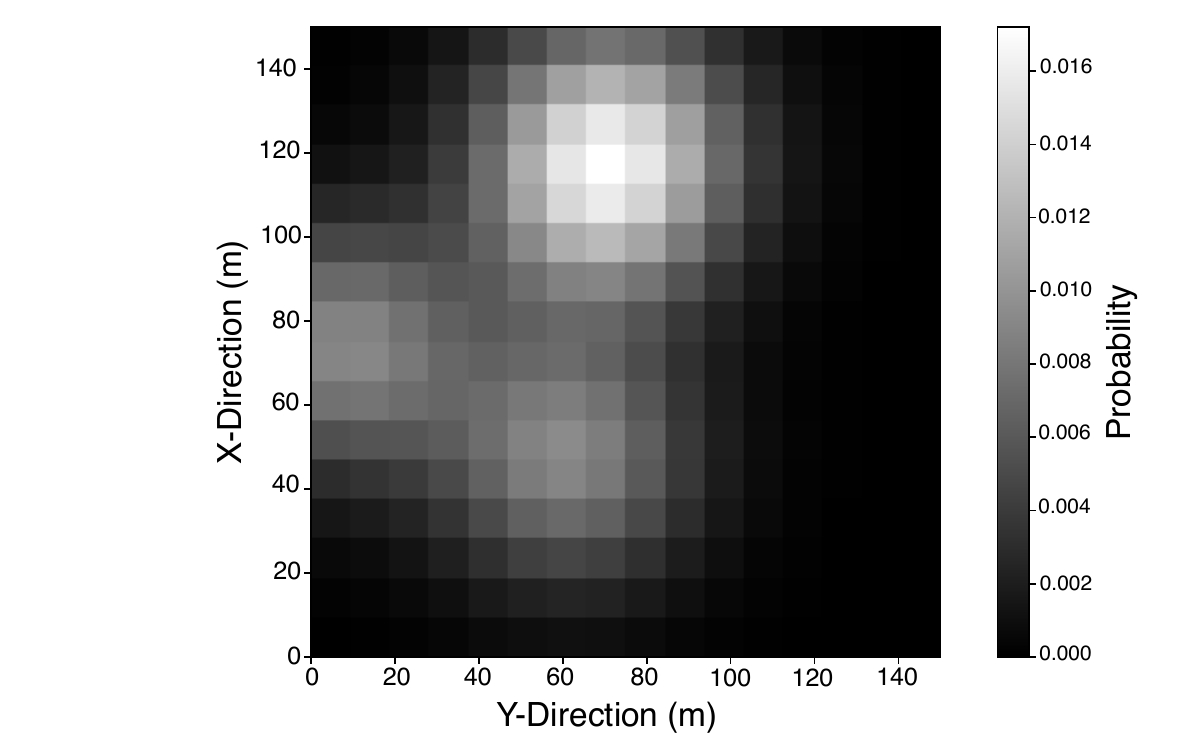}
    \caption{A random PDM ($p(\vec x)$)} \label{fig:pdmTrial4}
\end{figure}

The PDM defines the probability of finding a POI at any given position in a continuous manner (Fig. \ref{fig:pdmTrial4}). 
This is modelled as the sum of $G$ bivariate Gaussians
\cite{lin_hierarchical_2014}
such that a point on map at coordinate
$\vec x \in \mathbb R^2$ 
has a probability of containing the POI, $p(\vec x)$, as shown in Equation (\ref{eq:probability}).
\begin{equation}
	p(\vec x) =
	\frac{1}{G} 
	\sum^G_{i=0}
	\frac
	{
		\exp{
			\left[
				-\frac
				{1}
				{2}
				(\vec x - \vec \mu_i )^T\vec\sigma^{-1}_i(\vec x - \vec \mu_i) 
			\right]
		}
	}
	{\sqrt{4\pi^2\det{\vec\sigma_i}}}
    \label{eq:probability}
\end{equation}
Here $\vec \mu_i$ and $\vec \sigma_i$ are the mean location and covariance matrix of the $i$th bivariate Gaussian respectively. For the purpose of this study, these values are randomly generated with $G=4$. In a real-world scenario, this can be generated using an algorithm such as J1 for wilderness search and rescue \cite{ewers_gis_2023}.

\section{MISSION PLANNING}\label{sec:missionPlanning}
In order to generate a trustworthy autonomous mission plan, target points are generated using PDM-based search planning.  This set of target points is split into individual segments, where a segment consists of two sequential target points. The coordination algorithm attempts to identify safe paths for each of the five rovers in each segment, where each path is planned using a variant of the Rapidly-exploring Random Tree (RRT) algorithm.
\subsection{Target Generation}
\newcommand{\LHCGWCONV}{LHC\_GW\_CONV}

PDM-based search planning differs from classical A to B path planning in that it aims to \textit{accumulate} the maximum probability, $p(\vec x)$, along a path. The search planning algorithm selected for this work is \LHCGWCONV \cite{lin_uav_2009}.
\LHCGWCONV \hspace{1mm} approaches the path planning problem by segmenting the search area into $N \times M$ cells, such that each cell is as large as the search footprint. 
In this study, the search footprint has a diameter of $5$m (i.e. each of the five rovers in the team has a $1$m search diameter). 

\LHCGWCONV \hspace{1mm} is based on the Local Hill Climbing (LHC) optimisation method, which
considers all eight cells around the current position and selects the cell with the highest
probability value. This cell is selected for the next step and the previous cell is marked as having
been completely seen with a value of $0$. This does not prohibit future traversing of this cell, but
strongly discourages it to prevent the scenario of having being stuck with nowhere to go resulting
in early termination.  In the case that multiple cells have equal values, a $3 \times 3$ normalized
box blur convolution kernel, $\omega$, is applied to each equally valued cell (Equation (\ref{eq:convMatrix})).
\begin{equation}
    \omega = \frac{1}{9}\begin{bmatrix}
        1 & 1 & 1 \\
        1 & 1 & 1 \\
        1 & 1 & 1
    \end{bmatrix}
    \label{eq:convMatrix}
\end{equation}  
The convoluted value $p^{conv}(\vec x)$ for any position is found using Equation (\ref{eq:convVal}). 
\begin{equation}
    p^{conv}(\vec x) = \omega * p(\vec x) = 
    \sum_{i=-1}^{1}\sum_{j=-1}^{1}
    \omega(i,j)p\left(
        \vec x - 
        \begin{bmatrix}
            i\\j
        \end{bmatrix}
    \right)
    \label{eq:convVal}
\end{equation}
The cell with the largest vale of $p^{conv}(\vec x)$ is selected for the next step. Without any
further modifications, the algorithm would fully explore the nearest local maxima fully before
considering others. This is a common problem with LHC. To encourage exploring the entire PDM, the
concept of \textit{global warming}, GW, is introduced. Here, a value $C$ is subtracted from the PDM
a $l$ number of times, where $C=p_{max}/l$ and $p_{max}$ is the global maxima. The PDM is then
updated through Equation (\ref{eq:probGW}). 

\begin{equation}
    p'(\vec x)
    =
    \begin{cases}
        p(\vec x) - C,& p(\vec x) > C \\
        0, & else
    \end{cases}
    \label{eq:probGW}
\end{equation}

After all $l$ GW steps are completed, each path is evaluated against the original PDM $p(\vec x)$ and the one with the maximum accumulated probability is returned.

\subsection{Path Planning}
In this work, an RRT* path planning algorithm is implemented. The RRT algorithm was first set out by LaValle \cite{LaValle1998}. An RRT is a randomised data structure which facilitates path planning for non-holonomic vehicles. Karaman and Frazzoli \cite{karaman2011} introduced RRT*, an asymptotically optimal extension of RRT, i.e. as the number of nodes on the tree increases, the cost of the returned solution converges on an optimal region.

In the general form of RRT*, path cost is based purely on distance. However, for robots in 3D environments with varying terrain, the shortest path may not always be the preferred path. For planetary exploration robots, paths which are longer, but smoother and flatter are often preferable for robot safety. Equation (\ref{eqn:3dRrtCost}) shows the cost function implemented in order to produce paths which are obstacle-free, smooth, and flat \cite{takemura2017}. Four cost components are utilised: path length ($R$), roll ($\phi$), pitch ($\theta$), and required turning angle from the previous node to current node ($\Delta \psi$). 
\begin{equation}
    cost(q_i ) = W_R\dfrac{R_i}{N_R} +  W_{\phi}\dfrac{\phi_i}{N_\phi} +  W_{\theta}\dfrac{\theta_i}{N_{\theta}} +  W_{\psi}\dfrac{\Delta \psi_i}{N_\psi} 
    \label{eqn:3dRrtCost}
\end{equation}

In the above equation, $q_i$  is the node currently being checked, $W$ represents a weighting factor for each cost component such that the weights sum to $1$. The weights $W_{\phi}$ and $W_{\theta}$ are set to 0.4 with $W_{R}$ and $W_{\psi}$ set to $0.1$ such that the flatness of paths is prioritised. $N$ represents a normalisation factor to make each index dimensionless. These values are based on the maximum valid value of the respective cost components (i.e. the maximum step the RRT can take is $1$m, hence $N_R$ = $1$m). The respective normalisation factors are $N_R$=$1$m, $N_{\phi}$=$15^{\circ}$, $N_{\theta}$=$15^{\circ}$, and $N_{\psi}$=$60^{\circ}$.

For any given pair of start and target points, boundaries are set to ensure the RRT* path planner searches only a small, relevant chunk of the full mission site (i.e. $2$m clearance of the start and target points in both the X and Y directions). 
The RRT* planner searches the bounded area by growing the tree until the maximum number of nodes has been reached. The maximum number of nodes selected in this work is $1250$, which is sufficient to thoroughly search the bounded area, but does not incur a high run time. 

\subsection{Coordination}
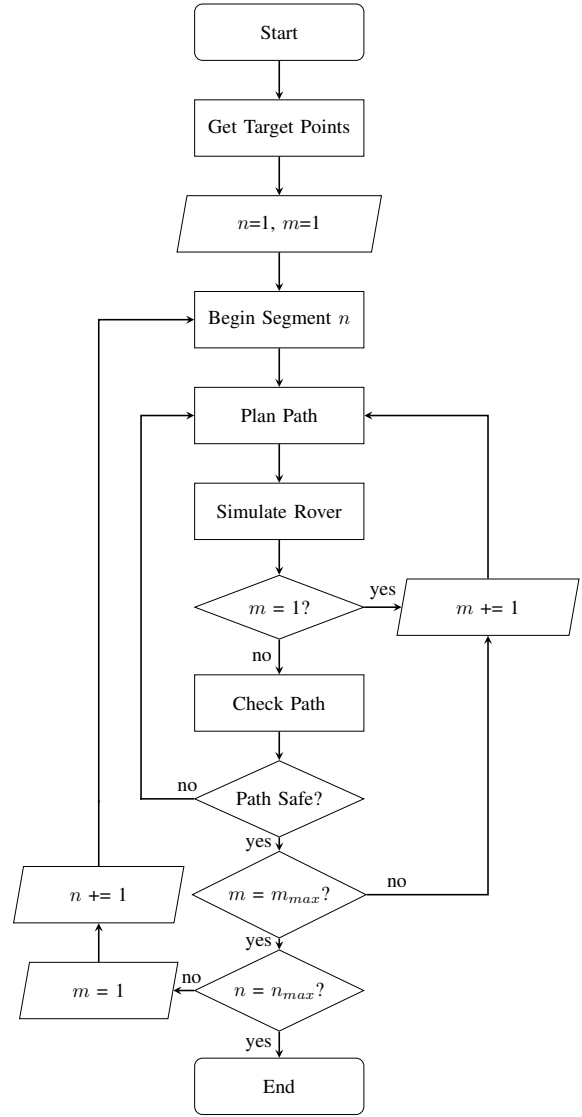
\begin{figure}[h]
    \centering
    \scalebox{0.75}{
    \begin{tikzpicture}[node distance=1.7cm]
        \node (start) [startstop] {Start};
        \node (pro0) [process, below of=start] {Get Target Points};
        \node (in1) [io, below of=pro0] {$n$=1, $m$=1};
        \node (pro1) [process, below of=in1] {Begin Segment $n$};
        \node (pro2) [process, below of=pro1] {Plan Path};
        \node (pro3) [process, below of=pro2] {Simulate Rover};
        \node (dec1) [decision, below of=pro3] {$m$ = 1?};
        \node (pro4) [process, below of=dec1] {Check Path};
        \node (dec2) [decision, below of=pro4] {Path Safe?};
        \node (dec3) [decision, below of=dec2] {$m$ = $m_{max}$?};
        \node (dec4) [decision, below of=dec3] {$n$ = $n_{max}$?};
        \node (end) [startstop, below of=dec4] {End};

        \node (in2) [io, right of=dec1, xshift = 2cm] {$m$ += 1};
        \node[left of=dec2, xshift = -0.8cm]  (pointA) {};    
        \node[left of=dec2, xshift = -0.89cm]  (pointB) {};    
        \node[left of=dec2, xshift = -1.5cm,yshift = -0.2cm] (pointC) {};  
        \node[left of=dec2, xshift = -1.5cm,yshift = 0.11cm]  (pointD) {};  
        \node[left of=dec2, xshift = -0.75cm,yshift = -0.14cm]  (pointE) {};  
        \node (in3) [io, left of=dec3, xshift = -1.5cm] {$n$ += 1};  
        \node (in4) [io, left of=dec4, xshift = -1.5cm] {$m$ = 1};  

        \draw [arrow] (start) -- (pro0);
        \draw [arrow] (pro0) -- (in1);
        \draw [arrow]  (in1) -- (pro1);
        \draw [arrow] (pro1) -- (pro2);
        \draw [arrow] (pro2) -- (pro3);
        \draw [arrow] (pro3) -- (dec1);
        \draw [arrow] (dec1) -- node[left] {no} (pro4);
        \draw [arrow] (pro4) -- (dec2);
        \draw [arrow] (dec2) -- node[left] {yes} (dec3);
        \draw [arrow] (dec3) -- node[left] {yes} (dec4);
        \draw [arrow] (dec4) -- node[left] {yes} (end);

        \draw [arrow] (dec1) -- node[above] {yes} (in2);         
        \draw [arrow] (in2) |- (pro2);   
        \draw [arrow] (dec3) -| node[above, very near start] {no} (in2);  
        \draw [thick] (dec2) -- node[above, very near start] {no} (pointB); 
        \draw [arrow] (pointE) |- (pro2);      
        \draw [arrow] (dec4) -- node[above, very near start] {no} (in4);  
        \draw [arrow] (in4) --  (in3); 
        \draw [thick] (in3) -- (pointD); 
        \draw [arrow] (pointC) |- (pro1); 
    \end{tikzpicture}
    } 
    \vspace*{3mm}
    \caption{Prioritised planning coordination flowchart}
    \label{fig:flowchart}
\end{figure}

The paths generated for each rover must be coordinated such that no collisions occur. For this purpose, prioritised planning is utilised. Prioritised planning is a 4D coordination methodology, which has been shown to eliminate dynamic collisions under nominal conditions (i.e. when no faults are present in the system), outperforming other common coordination algorithms such as fixed path coordination \cite{swinton2022novel}. Using this method, an initial set of safe paths can be coordinated offline. Each rover is awarded a priority index. The highest priority rover’s path is planned first, and a simulation is run to acquire 4D positional and temporal data as the rover traverses its planned path. The algorithm then attempts to plan a path for the second rover, comparing the positions of both rovers at each time-step to check for potential collisions. If collisions are detected, another path planning attempt is made for the second rover. This process repeats until the second rover’s path is deemed ‘safe’. The algorithm then attempts plan a collision free path for each subsequent rover, descending in priority. 
This process is outlined in Fig. \ref{fig:flowchart}, where $n$ is the segment index (with a maximum value of n\textsubscript{max}), and $m$ is the rover index. The maximum rover index is $m_{max}$, which has a value of 5 in this case. 

    \vspace*{-3mm}
\section{RESULTS}\label{sec:results}
To evaluate the ability of the multi-layer mapping and mission planning methodologies to generate safe and efficient autonomous paths, various test scenarios have been considered in the Jezero crater mission site. A representative scenario has been illustrated where appropriate. In each test scenario, the PDM is varied randomly. This results in unique target generation for each test case, and consequently different areas of the mission site are explored. The experimental results are set out as follows. First, the quality of the paths generated by the mission planner is evaluated. Second, the safety of the coordinated paths is assessed. Finally, the efficiency of autonomous exploration plan is analysed. 

\subsection{Generation of Safe Paths: Single Rover}
\begin{figure}[htbp]
    \centering
     \includegraphics[clip, trim=3cm 0cm 0.9cm 0cm, width=0.9\linewidth]{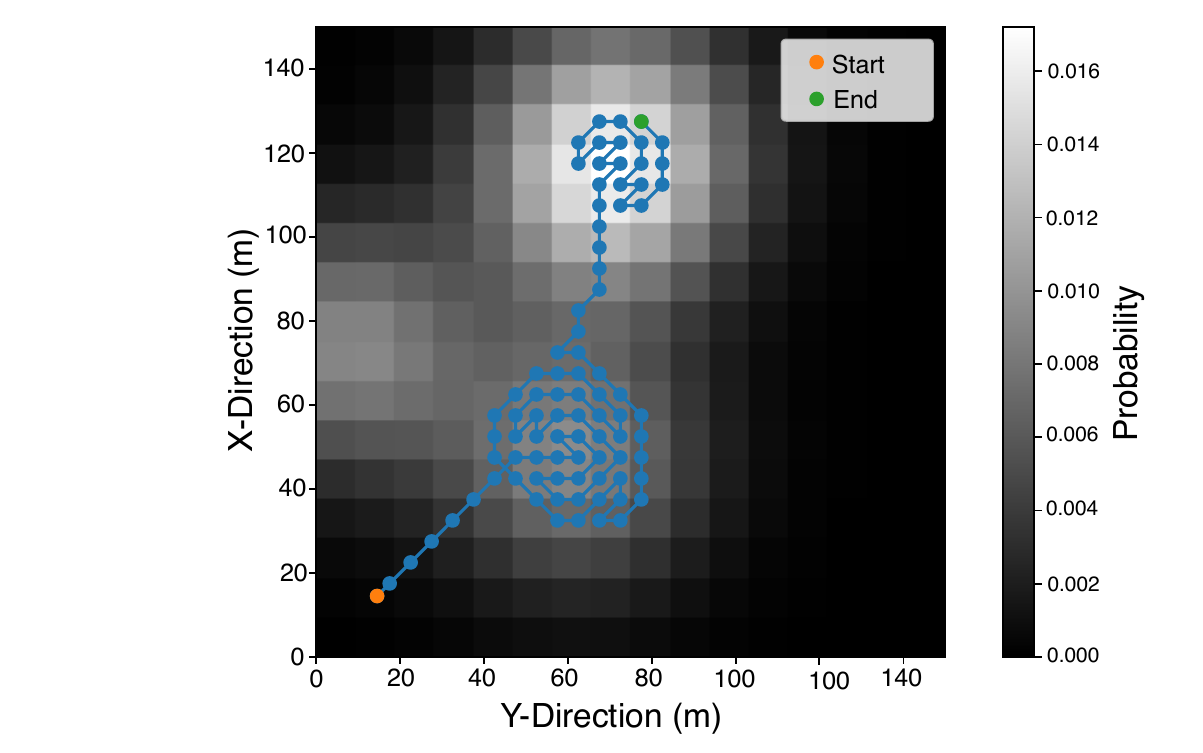}
    \caption{A set of target points generated using a random PDM ($p(x)$) and \LHCGWCONV } \label{fig:missionPlanningTrial105}
\end{figure}

In each mission scenario, a random PDM is generated, and a corresponding set of targets is evaluated using \LHCGWCONV. This set of targets maximises the accumulated probability of capturing the POI along the path.  Fig. \ref{fig:missionPlanningTrial105} shows the set of target points generated for a representative scenario. In the case shown, it can be seen that \LHCGWCONV guides the mission planner away from searching only one local maxima.

\vspace*{-3mm}

\begin{figure}[htbp]
    \centering
     \includegraphics[clip, trim=1cm 0cm 1cm 1cm, width=0.9\linewidth]{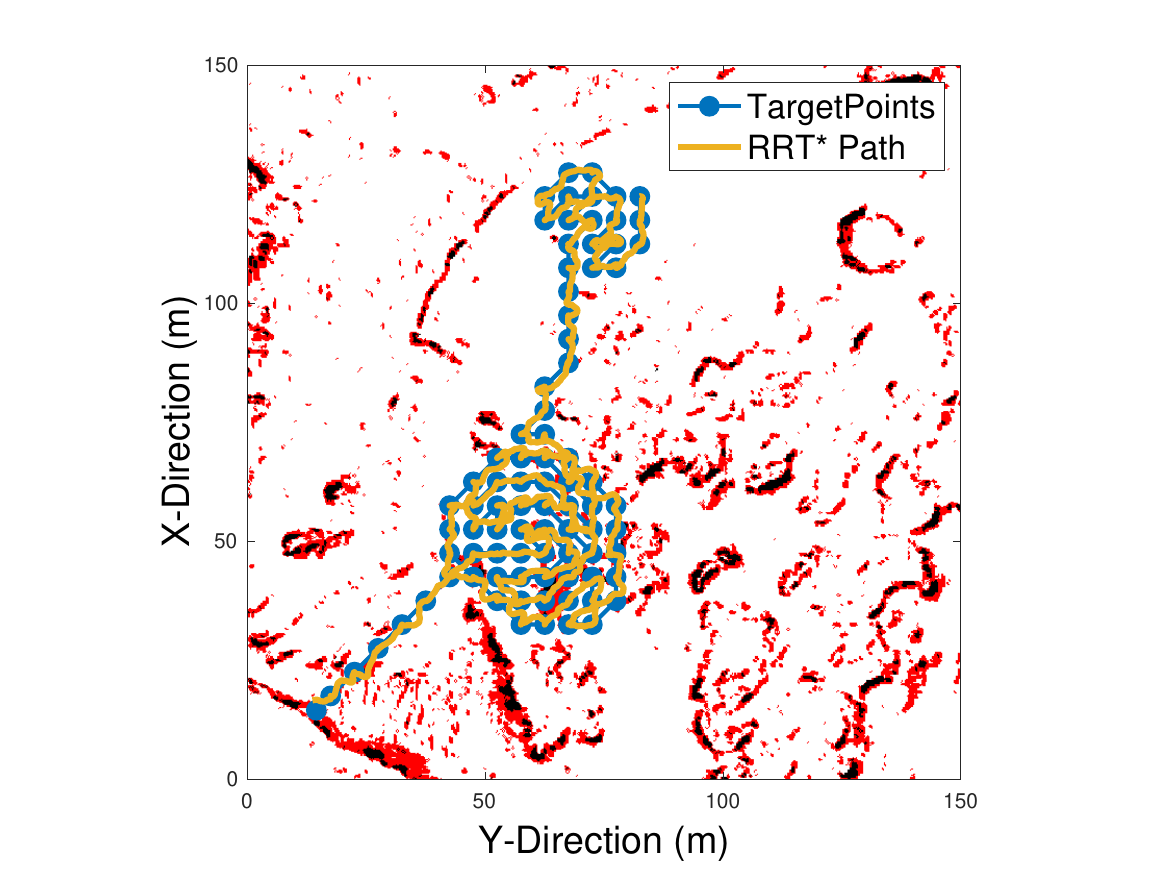}
    \caption{A representative example of a path generated by RRT* to explore each of the \LHCGWCONV \hspace{1mm}targets} \label{fig:rrtPath105}
\end{figure}

\vspace*{-3mm}
For each pair of sequential targets, the RRT* path planner attempts to find a safe path using the traversability map. Fig. \ref{fig:rrtPath105} shows a path generated over a full set of target points.

Data has been sampled at a frequency of $10$Hz, resulting in a total of $942321$ samples for the rover poses. 
Throughout the test set, rover pitch and roll measurements exceeded the nominal threshold of $15^{\circ}$ on only $25$ occasions. This means that the RRT* path planner has been able to provide paths that keep the rovers within their nominal pitch and roll limits for $99.99$\% of operation. The few instances where pitch or roll exceed nominal limits are due to slip as the rover attempts to traverse waypoints, causing it to veer slightly off the ideal trajectory.

\subsection{Generation of Safe Paths: Rover Team}

\begin{figure}[htbp]
    \centering
     \includegraphics[clip, trim=3cm 0cm 0.9cm 0cm, width=0.9\linewidth]{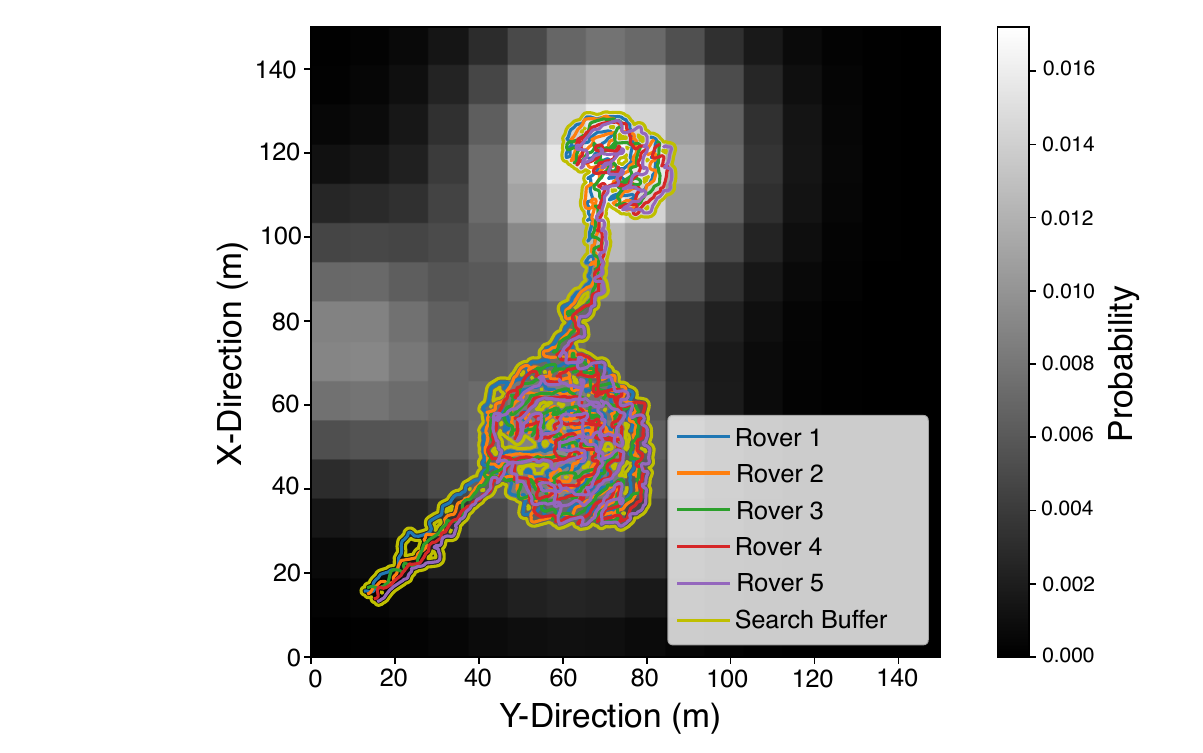} 
     \caption{A representative example of fully coordinated rover team trajectories over a full mission scenario}
     \label{fig:resultsTrajectories}
\end{figure}

The prioritised planning coordination algorithm invalidates any RRT* path planning attempts that cause a rover to collide with higher priority team mates. As such, the selected method generates safe paths in each mission scenario.  
Fig. \ref{fig:resultsTrajectories} shows an example of the full trajectories of five rovers in a test case. The boundaries of the team's sensor footprint are shown by the search buffer. Gaps in the search buffer can be observed where the rovers follow paths around impassable terrain. 
\vspace*{-5mm}
\subsection{Exploration Efficiency}
The full mission site covers $22500$m$^2$. 
Considering a team of five rovers, each with a sensor footprint of 1m radius, each rover would be required to travel $\sim4500$m for the team to acquire an accumulated probability of locating the POI approaching 1. 

\begin{figure}[htbp]
    \centering
     \includegraphics[width=0.85\linewidth]{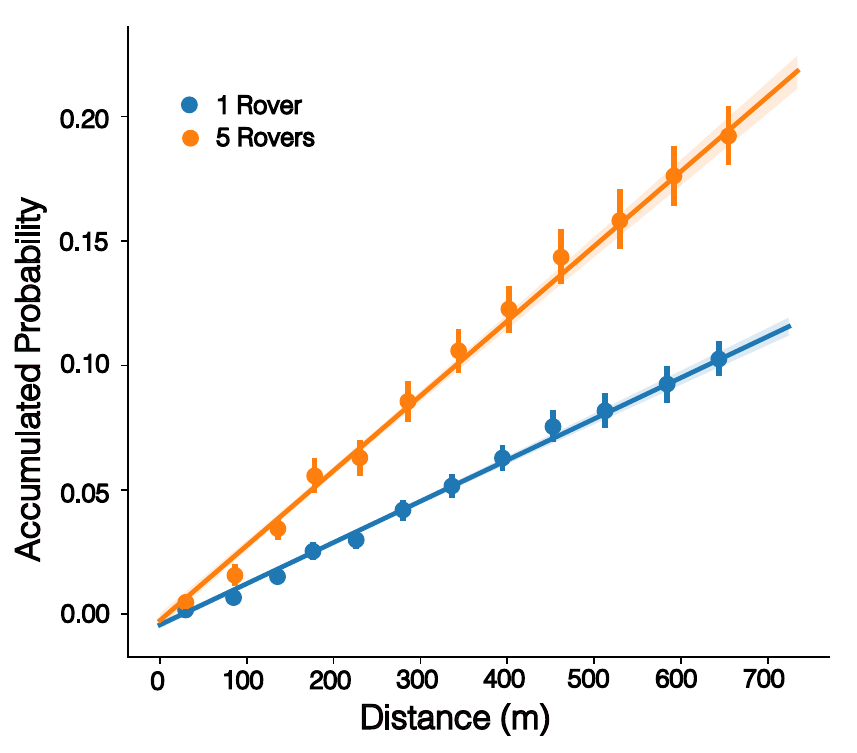} 
     \caption{Accumulated probability over distance travelled for a single rover compared to a team of five rovers}
     \label{fig:singleVsTeam}
\end{figure}

Over the mission scenarios considered, the average time taken for a rover team to traverse the 64 target points was 2295.25 seconds (38.26 minutes), with an average trajectory of 653.68m. 
By utilising the proposed autonomous exploration method, the rovers are able to acquire almost $p(x)$ = 0.2  while travelling only 650m. 
Fig. \ref{fig:singleVsTeam} shows the increase in accumulated probability for a single rover, compared to that of a team of five rovers. 

These results show that the proposed methodologies allow a large area to be searched effectively by a small team of rovers, generating paths which increase accumulated probability and can be carried out in less than one sol of operation. 

\section{CONCLUSIONS} \label{sec:conclusions}

This work tackled the problem of autonomous exploration for multi-rover teams by considering a mapping approach to identify high-interest, safe, regions of terrain, and a novel mission planning methodology that allows the rover team to safely and efficiently explore a given mission region. 
The mapping approach in this work was composed of three stages: a 3D digital terrain model of Jezero crater (generated using HiRISE orbiter data), a traversability analysis of the mission site, and a PDM which maps the areas of the mission site most likely to contain scientific POI. 
The mission planning methodology consisted of three stages. First, PDM-based search planning was used to find a set of targets which accumulate the highest probability of finding scientific POI. Second, a 4D RRT* path planner was used to identify flat, smooth paths between targets. Finally, prioritised planning coordination was used to ensure that each rover path was safe i.e. did not incur any collisions. 

The performance of the proposed methods has been evaluated over a set of randomly generated PDMs in the mission site of Jezero crater, Mars. It has been shown that collision-free autonomous exploration can be carried out efficiently over an area of $22500$m$^2$ in an average time of 38.26 minutes. Further, the rover trajectories generated during testing had an average length of 653.68 metres; comparable to the current record for longest distance driven without human review by a planetary exploration rover, and significantly exceeding the longest distance driven on a single sol. Therefore, the approach proposed in this paper successfully enables safe and efficient autonomous exploration of a 3D environment using a team of planetary exploration rovers.

\addtolength{\textheight}{-12cm}   










\begin{thebibliography}{99}

\bibitem{farley2020} Farley KA, Williford KH, Stack KM, Bhartia R, Chen A, de la Torre M, Hand K, Goreva Y, Herd CD, Hueso R, Liu Y. ``Mars 2020 mission overview". Space Science Reviews. 2020 Dec;216:1-41.

\bibitem{fong2021} Fong T, Allen BD, Azimi S. ``Autonomous Systems \& Robotics for Lunar Surface Infrastructure''. In ASCEND. 2021 Nov.

\bibitem{swinton2022novel} Swinton S, McGookin E. ``A Novel, RRT* Based Approach to the Coordination of Multiple Planetary Rovers". In 2022 UKACC 13th International Conference on Control (CONTROL) 2022 Apr 20 (pp. 88-93). IEEE.

\bibitem{fink2015}W. Fink W, Baker VR, Schulze-Makuch D, Hamilton CW, Tarbell MA, ``Autonomous exploration of planetary lava tubes using a multi-rover framework". In 2015 IEEE Aerospace Conference, 2015, pp. 1-9, doi: 10.1109/AERO.2015.7119315.

\bibitem{vaquero2018} Vaquero T, Troesch M, Chien S. ``An approach for autonomous multi-rover collaboration for mars cave exploration: Preliminary results". In International Symposium on Artificial Intelligence, Robotics, and Automation in Space (i-SAIRAS 2018). Also appears at the ICAPS PlanRob 2018.


\bibitem{stOnge2019} St-Onge D, Kaufmann M, Panerati J, Ramtoula B, Cao Y, Coffey EB, Beltrame G. ``Planetary exploration with robot teams: Implementing higher autonomy with swarm intelligence". IEEE Robotics \& Automation Magazine. 2019 Nov 12;27(2):159-68.

\bibitem{bajracharya2008} Bajracharya M, Maimone MW, Helmick D. ``Autonomy for mars rovers: Past, present, and future". Computer. 2008 Dec 12;41(12):44-50.

\bibitem{verma2023} Verma V, Maimone MW, Gaines DM, Francis R, Estlin TA, Kuhn SR, Rabideau GR, Chien SA, McHenry MM, Graser EJ, Rankin AL. ``Autonomous robotics is driving Perseverance rover’s progress on Mars". Science Robotics. 2023 Jul 26;8(80):eadi3099.

\bibitem{planetarySociety} The Planetary Society. ``The Planetary Exploration Budget Dataset". Available from: https://www.planetary.org/space-policy/planetary-exploration-budget-dataset [Accessed: 2024/01/13] 

\bibitem{servoCityRBR} ServoCity. Bogie Runt Rover [Internet]. [Cited: 2024 April 08]. Available from: \url{https://www.servocity.com/bogie-runt-rover/}

\bibitem{flessa2014} Flessa T, McGookin EW, Thomson DG. ``Taxonomy, systems review and performance metrics of planetary exploration rovers". In2014 13th International Conference on Control Automation Robotics \& Vision (ICARCV) 2014 Dec 10 (pp. 1554-1559). IEEE.

\bibitem{fossen1994}Fossen T.I. Guidance and control of Ocean Vehicles. \textit{Wiley \& Sons}. 1994

\bibitem{zurek2007} Zurek RW, Smrekar SE. ``An overview of the Mars Reconnaissance Orbiter (MRO) science mission". Journal of Geophysical Research: Planets. 2007 May;112(E5).

\bibitem{hiriseSpecs2012} HiRISE. University of Arizona. ``Camera Technical Specifications". 2012. Available from: https://www.uahirise.org/specs/ [Accessed: 2022/12/07]


\bibitem{rankin2020} Rankin A, Maimone M, Biesiadecki J, Patel N, Levine D, Toupet O. Driving curiosity: Mars rover mobility trends during the first seven years. In2020 IEEE Aerospace Conference 2020 Mar 7 (pp. 1-19). IEEE.

\bibitem{lin_hierarchical_2014} Lin L, Goodrich MA, ‘Hierarchical Heuristic Search Using a Gaussian Mixture Model for UAV Coverage Planning’, IEEE Trans. Cybern., vol. 44, no. 12, pp. 2532–2544, Dec. 2014, doi: 10.1109/TCYB.2014.2309898.

\bibitem{ewers_gis_2023} Ewers JH, Anderson D, Thomson D, ‘GIS Data Driven Probability Map Generation for Search and Rescue Using Agents’, in IFAC World Congress 2023, 2023, pp. 1466–1471. doi: 10.1016/j.ifacol.2023.10.1834.

\bibitem{lin_uav_2009} Lin L, Goodrich MA, ‘UAV intelligent path planning for wilderness search and rescue’, 2009 IEEE/RSJ International Conference on Intelligent Robots and Systems, IROS 2009, vol. 0, no. 1, pp. 709–714, 2009, doi: 10.1109/IROS.2009.5354455.

\bibitem{LaValle1998}LaValle S. Rapidly-exploring random trees: A new tool for path planning. Research Report 9811. 1998.

\bibitem{karaman2011} Karaman S, Frazzoli E. Sampling-based algorithms for optimal motion planning. The international journal of robotics research. 2011 Jun;30(7):846-94.

\bibitem{takemura2017} Takemura R, Ishigami G. ``Traversability-based RRT* for planetary rover path planning in rough terrain with LIDAR point cloud data". Journal of Robotics and Mechatronics. 2017 Oct 20;29(5):838-46.


\end{thebibliography}
\end{document}